\documentclass[runningheads]{llncs}
\usepackage{graphicx}
\graphicspath{ {./images/} }
\usepackage{floatrow}
\newfloatcommand{capbtabbox}{table}[][\FBwidth]
\usepackage{dirtytalk}

\usepackage{multirow}
\usepackage{tikz}

\begin{document}

\title{Entity Type Prediction in Knowledge Graphs using Embeddings}
%
%
\author{Russa Biswas\inst{1,2} \and Radina Sofronova \inst{2} \and  Mehwish Alam\inst{1,2} \and
Harald Sack\inst{1,2}}
\authorrunning{Russa Biswas et al.} 
%
%
\institute{FIZ Karlsruhe -- Leibniz Institute for Information Infrastructure, Germany \\
\and
Karlsruhe Institute of Technology, Institute AIFB, Germany \\
\email{radina.sofronova@student.kit.edu},\\
\email{\{russa.biswas, mehwish.alam, harald.sack\}@fiz-karlsruhe.de}}
\maketitle              
\begin{abstract}
Open Knowledge Graphs (such as DBpedia, Wikidata, YAGO) have been recognized as the backbone of 
diverse applications in the field of data mining and information retrieval. Hence, the completeness and correctness of the Knowledge Graphs (KGs) is vital. Most of these KGs are mostly created either via an automated information extraction from Wikipedia snapshots or information accumulation provided by the users or using heuristics. However, it has been observed that the type information of these KGs is often noisy, incomplete and incorrect. To deal with this problem a multi-label classification approach is proposed in this work for entity typing using KG embeddings. We compare our approach with the current state-of-the-art type prediction method and report on experiments with the KGs.

\end{abstract}
%
%
%
\section{Introduction}
Open Knowledge Graphs (KGs) such as DBpedia, Wikidata, YAGO, etc. have been recognized as the foundations for diverse KG based applications including Natural Language Processing, data mining and Information Retrieval. Most of these KGs are created either via automated information extraction from Wikipedia snapshots, information accumulation provided by the users or by using heuristics. However, each KG follows a different knowledge organization and is based on differently structured ontologies. Moreover, it has been observed that type information are often noisy or incomplete. On the other hand, these KGs contain huge amount of data which makes it difficult to be used by the applications. Therefore, recent years have witnessed an extensive research on the latent representation of the KGs in a low dimensional vector space. In this work, the proposed method addresses the entity typing problem in DBpedia using the embeddings as a multi-label classification problem. 

\begin{figure}[t!]
\begin{floatrow}
\ffigbox{
\includegraphics[width=0.5\textwidth]{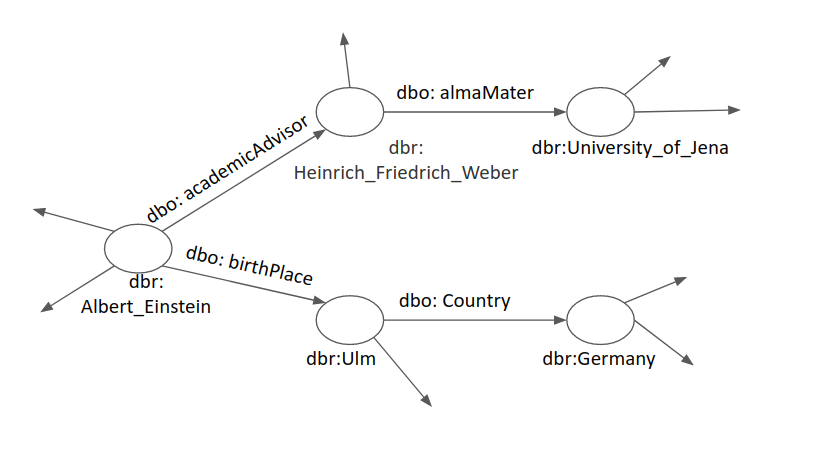}}
{\caption{Exempt of a subgraph from DBpedia}
\label{fig:intro}}
\capbtabbox{%
  \begin{tabular}{ p{1.8cm}p{1.5cm}p{3cm} }
 \hline
 Class & \#Entities & \#Coarse-Grained entities\\
 \hline
    SportsTeam & 352006 & 320835 (91.1\%)\\
    Company & 70208 & 55524 (79.1\%)\\
    Settlement & 478906 & 246163 (51.4\%) \\
    Activity & 19464 & 8824 (45.3\%)\\
    Event & 76029 & 19418 (25.5\%)\\
 \hline
\end{tabular}
}{%
  \caption{Distribution of entities in subclasses}%
  \label{table:intro}
}
\end{floatrow}
\end{figure}

Entity typing is the process of assigning a type to an entity and is a fundamental task in KG completion. For example, the triple {\tt <\textbf{\texttt{dbr:Albert\_Einstein}, \texttt{rdf:type}, \texttt{dbo:Scientist}}>} states that Albert Einstein is assigned to the type class Scientist. The type information in DBpedia is derived directly by an external extraction framework from the Wikipedia infobox types. Since, the Wikipedia is a crowd sourced encyclopedia, hence this type information is often incomplete. Therefore, a huge number of entities in DBpedia are assigned to a coarse grained \texttt{rdf:type}. Table~\ref{table:intro} provides the distribution of entities of five types. For e.g.,  class \textit{dbo:SportsTeam} has 14 subclasses in DBpedia and 352006 entities, out of which only 8.9\% are assigned to its subclasses. Hence, there arouses a necessity to have fine grained types for the entities in the KGs.

On the other hand, most of the existing state-of-the-art KG embedding approaches such as translational approaches such as TransE~\cite{bordes2013translating}, TransR~\cite{lin2015learning}, etc. exploit only the structure of the KG.
However, besides the structural information, implicit textual semantic information is also stored in the KGs as illustrated in Figure~\ref{fig:intro}. This subgraph depicts, the \textbf{birthplace} of \say{Albert Einstein} is \say{Ulm}, which is located in the \textbf{country} \say{Germany}. The labels of the triples in the subgraph, such as birthplace, country, Ulm, etc. contains implicit textual information in the graph, that is not captured in translational embedding models. 

In this paper, a multi-label classification approach is proposed for fine grained entity typing. To do so, the model uses different existing word embedding models such as Word2Vec~\cite{DBLP:journals/corr/abs-1301-3781}, GloVe~\cite{pennington2014glove}, and FastText~\cite{bojanowski2017enriching} to learn the KG embeddings capturing the graph structure as well as the implicit textual information available. The main contributions of this paper are:
\begin{itemize}
    \item Vector representation of entities and relations in DBpedia using the existing word embedding models.
    \item A multi-label classification based approach for fine grained entity typing. 
    \item An analysis and comparison of the aforementioned word embedding models for the task of entity type prediction.
\end{itemize}
The rest of the paper is structured as follows. To begin with, a review of the related work is provided in Section~2. Section~3 accommodates the detailed description of the approach followed by experimental setup and report on the results in Section~4. Finally, an outlook of future work is provided in Section~5.

\vspace{-.6cm}

\section{Related Work}
This section presents the prior related works on entity typing considering both Wikipedia Infobox type prediction as well as RDF type prediction.

\noindent\textbf{\textit{Wikipedia Infobox Type Prediction.}}
One of the initial works in this domain was proposed by Wu et al.\cite{DBLP:conf/cikm/WuW07}. To do so, KYLIN considers pages having similar infoboxes, determines the common attributes in them to learn a CRF extractor. Sultana et al.\cite{DBLP:conf/cikm/SultanaHBDRDL12} focuses on automated approach by training a SVM classifier on the feature set of TF-IDF on the first \textit{k} sentences of an article as well as on categories and Named Entity mentions. Biswas et al.\cite{biswas2018wikipedia,biswas2018predicting} provides a neural network based approach for infobox prediction using word embeddings on abstract, table of contents, and categories of Wikipedia articles.

\noindent\textbf{\textit{RDF Type Prediction.}}
A statistical heuristic link based type prediction mechanism, SDTyped, has been proposed by Paulheim et al. and was evaluated on DBpedia and OpenCyc \cite{paulheim2013type}. Another RDF type prediction of KGs has been studied by Melo et
al. \cite{melo2016type}, where the type prediction of the KGs is performed via the hierarchical SLCN algorithm using a set of incoming and outgoing relations as features for classification. In~\cite{DBLP:journals/ws/KliegrZ16}, the authors propose a supervised hierarchical SVM classification approach for DBpedia by exploiting the contents of Wikipedia articles. However, none of these methods exploit embeddings to perform the type prediction. In this work, different word embedding algorithms will be exploited on the KGs for the task of entity typing.

\section{Entity Typing using Embeddings}
The task of entity type prediction is multi-label classification problem considering the entity type information as classes which is discussed in this section.

\subsection{Word Embeddings on KGs}

Each triple or fact in the KG is considered as a sentence where the relation serves as a verb and the two entities are considered as the subject and the object of this relation in the sentence. For e.g., $<dbr:Albert\_Einstein, dbo:birthplace, dbr:Ulm>$ is considered as a sentence 
These sentences are then used as a corpus for all the three word embeddings. The URIs are considered for training. The dimension of the vectors for each of the embedding models is 100 and the embeddings from all the models for DBpedia available in our GitHub~\cite{gitlink}


\noindent\textit{\textbf{Word2Vec. }}It aims to learn the distributed representation for words reducing the high dimensional word representations in large corpus. It comprises of two model architectures, Continuous Bag of Words (CBOW) and Skip-gram. In CBOW approach, the model predicts the current word from a window of context words. On the other hand, the skip-gram model tries to predict the context words based on the current word. In this work, the CBOW approach of Word2Vec model has been used to learn the vector representation of the entities and relations in the KG based on the context entity or relation.  

\noindent\textit{\textbf{FastText. }}FastText is an extension of the word2vec model, which follows both CBOW and Skip-gram architectures. The main difference with the Word2Vec is that it learns the representation of each word in the corpus as n-gram characters. This benefits in capturing representations for shorter or rare words which can be obtained by breaking down words into n-grams to get its embeddings. Therefore, it would help in having embeddings for unseen facts in KGs.

\noindent\textit{\textbf{GloVe. }}GloVe is another word embedding model which exploits the global word-word co-occurrence statistics in the corpus. The model is essentially a log-bilinear model with a weighted least-squares objective. The main underlying intuition is that ratios of word-word co-occurrence probabilities have the potential for encoding some form of meaning. The co-occurrence of the entities and the properties is important in learning the latent representation of KGs. 

\subsection{Entity Typing}

Two approaches have been used to determine the entity types in this work, (i) a supervised Convolutional Neural Network (CNN) based approach and (ii) vector similarity. 

\noindent\textit{\textbf{Convolutional Neural Network. }}The entity typing problem is converted to a classification problem with the \texttt{rdf:type} as classes in which, a 1D CNN model~\cite{kim2014convolutional} built on top of the embedding models. The model takes into account the vectors of the entities generated from the embedding models and predicts its type. The model consists of a convolutional layer which involves a feature detector followed by a global max pool layer. The activation function \texttt{ReLu} has been used in the convolutional layer. The output of the pooling layer is used which is then passed through a fully connected final layer, in which the sigmoid function calculates the probabilities of an entity belonging to different classes. The filter size taken is 128, with kernel sizes ${3,4,6}$, are chosen for the model.

\noindent\textit{\textbf{Vector Similarity. }}In order to assign fine-grained type to an entity with an already assigned coarse-grained type, class hierarchy in DBpedia has been exploited. For e.g., in DBpedia, for the entity \textit{dbr:Baker\&McKenzie}, the \texttt{rdf:type} class is \textit{dbo:LawFirm}. Next, class hierarchy of \textit{dbo:LawFirm} is traversed to find the highest level parent class \textit{dbo:Organisation} after \textit{dbo:Agent}. Now, all the subclasses of  \textit{dbo:Organisation} in the hierarchy is extracted and the cosine similarity between all the subclasses and the entity \textit{dbr:Baker\&McKenzie} has been calculated. Since the entities of a class represent the characteristic features of the class, the average vector of the entity vectors belonging to a certain class has been chosen as the class vector.


\section{Experiments and Results}
This section contains description of the experiments and analysis of the results. 

\noindent\textit{\textbf{Dataset. }}In order to have fine-grained type prediction of the entities which are already coarse-grained typed in DBpedia 2016-10\footnote{\url{https://wiki.dbpedia.org/downloads-2016-10}}, 3 datasets have been generated to evaluate the method. To determine the robustness of the method, the datasets comprise of classes with less number of entities as well as the ones with large entity count. The statistics of the dataset is provided in Table~\ref{results}.

\noindent\textit{\textbf{Results. }} The vector similarity approach is considered as the baseline model in this work. The CNN model is evaluated on 80\%-20\% of training and test split of each of the dataset as depicted in Table~\ref{results}. It is trained with a batch size of 32, 125 hidden layers and 1000 epochs.

\begin{table}[t!]
\begin{tabular}{cccccccccc}
\hline
\multirow{4}{*}{Datasets} & \multicolumn{9}{c}{Models (Results in Accuracy)} 
\\ \cline{2-10} 
 & \multicolumn{3}{c}{Word2Vec} & \multicolumn{3}{c}{FastText} & \multicolumn{3}{c}{GloVe}  \\ \cline{2-10}  & \multicolumn{2}{c}{\begin{tabular}[c]{@{}c@{}}Vector 
 \\ Similarity\end{tabular}} & \multirow{2}{*}{CNN} & \multicolumn{2}{c}{\begin{tabular}[c]{@{}c@{}}Vector 
 \\ Similarity\end{tabular}} & \multirow{2}{*}{CNN} & \multicolumn{2}{c}{\begin{tabular}[c]{@{}c@{}}Vector 
 \\ Similarity\end{tabular}} & \multirow{2}{*}{CNN} 
 \\ \cline{2-3} \cline{5-6} \cline{8-9}
    & Hits@3   & Hits@1 &  & Hits@3 & Hits@1  &  & Hits@3  & Hits@1  &   \\ \hline
\begin{tabular}[c]{@{}c@{}}59 classes,
\\ 500 entities/class\end{tabular} & 47.83\%   & 28.46\%   & \textbf{56\%}      & 29.81\%                   & 17.44\%  & \textbf{54\%} & 7.07\%  & 3.54\%    & \textbf{53.7\%}        \\ \hline
\begin{tabular}[c]{@{}c@{}}86 classes,
\\ 2k entities/class\end{tabular} & 58\%   & 39.4\%   & \textbf{58.4\%}      & 43.81\%                   & 31.16\%  & \textbf{56\%} & 15.9\%  & 8.2\%    & \textbf{55\%}        \\ \hline
\begin{tabular}[c]{@{}c@{}}81 classes,\\ 4k entities/class\end{tabular} & 58\%                         & 39.7\%  & \textbf{62\%}        & 44.3\%   & 31.4\%       & \textbf{59\%} & 16.2\%    & 8.4\%           & \textbf{55.8\%}  \\ \hline
\end{tabular}
\caption{Experimental Results}
\label{results}
\end{table}

It has been observed from the results that the CNN built on the top of the embedding models achieved better results in the entity typing task. However, the vector similarity results with Hits@3 for the word2vec vectors is comparable to CNN for the 86 classes dataset. The results of vector similarity depict that the vectors generated by the GloVe model is not so similar to each other, even then the CNN predicts the correct type with a much better accuracy. 
Also, 81 classes is a subset of 86 classes, with more number of entities per class which strengthens the fact the neural network models work better with more data. For the dataset with 4000 entities per class, the CNN works the best for all the embedding models as compared to the other methods. 

Also, the method has been compared with the available SDTyped dataset\footnote{\url{http://downloads.dbpedia.org/2016-10/core-i18n/en/instance\_types\_sdtyped\_dbo\_en.ttl.bz2}}. This dataset consists of the entity types predicted by SDType method. It is to be noted that only a small fraction of entities are common between the SDType dataset and our dataset as depicted in column 3 of Table~\ref{table:sdtype}. The count of the entities in SDTypes whose type information matches the ground truth is provided in the last column of same table. Due to huge differences in the datasets, a direct comparison of the models with this dataset is not possible. However, an analysis based only on the overlapping entities is available in the GitHub~\cite{gitlink}.

\begin{table}[]
\centering

\begin{tabular}{cccc}
\hline
Datasets & \begin{tabular}[c]{@{}c@{}}\#Entities in \\our dataset (E)\end{tabular} & \begin{tabular}[c]{@{}c@{}}\#$E \cap \#E_{SDType}$\\  \end{tabular} & \begin{tabular}[c]{@{}c@{}}\#$E_{SDType}$\\ $= GroundTruth$\end{tabular} \\ \hline
\begin{tabular}[c]{@{}c@{}}59 classes,\\ 500 entities/class\end{tabular} & 28106 & 7425 (26\%) & 6115 \\ \hline
\begin{tabular}[c]{@{}c@{}}86 classes,\\ 2k entities/class\end{tabular} & 172000 & 57467 (33\%)& 46222 \\ \hline
\begin{tabular}[c]{@{}c@{}}81 classes,\\ 4k entities/class\end{tabular} & 324000 & 109948 (34\%)& 89301 \\ \hline
\end{tabular}%
\caption{Analysis with the SDType data}
\label{table:sdtype}

\end{table} 
\section{Conclusion and Future Work}
In this paper, different word embeddings approaches for entity typing in a KG have been analyzed. The achieved results demonstrate that vectors coupled with CNN works better for the task. On the other hand, set theory concept\footnote{A set is represented by its members, which exhibit the same properties.} when applied to generate the class vectors from the entity vectors proved to be beneficial. In future, these embedding models would be used for other KG completion tasks such as link prediction, triple classification, etc. Also, for the entity typing task, more information to be included in this embedding space such the DBpedia categories to improve the results.   

\small{
\bibliographystyle{splncs04}
\bibliography{bibliography}
}
\end{document}